\documentclass{ieeeaccess}
\usepackage{cite}
\usepackage{amsmath,amssymb,amsfonts}
\usepackage{algorithmic}
\usepackage{graphicx}
\usepackage{textcomp}
\usepackage{url}
\usepackage{subcaption}
\usepackage{hyperref}
\usepackage{bm}
\makeatletter
\AtBeginDocument{\DeclareMathVersion{bold}
\SetSymbolFont{operators}{bold}{T1}{times}{b}{n}
\SetSymbolFont{NewLetters}{bold}{T1}{times}{b}{it}
\SetMathAlphabet{\mathrm}{bold}{T1}{times}{b}{n}
\SetMathAlphabet{\mathit}{bold}{T1}{times}{b}{it}
\SetMathAlphabet{\mathbf}{bold}{T1}{times}{b}{n}
\SetMathAlphabet{\mathtt}{bold}{OT1}{pcr}{b}{n}
\SetSymbolFont{symbols}{bold}{OMS}{cmsy}{b}{n}
\renewcommand\boldmath{\@nomath\boldmath\mathversion{bold}}}
\makeatother

\def\BibTeX{{\rm B\kern-.05em{\sc i\kern-.025em b}\kern-.08em
    T\kern-.1667em\lower.7ex\hbox{E}\kern-.125emX}}

\begin{document}
\history{Date of publication xxxx 00, 0000, date of current version xxxx 00, 0000.}
\doi{10.1109/ACCESS.2024.0429000}

\title{Efficient Few-Shot Learning in Remote Sensing: Fusing Vision and Vision-Language Models}
\author{\uppercase{Jia Yun Chua}\authorrefmark{1},
\uppercase{Argyrios Zolotas}\authorrefmark{2}, and \uppercase{Miguel Arana-Catania}\authorrefmark{3}
}

\address[1]{Cranfield University (e-mail: jiayun.chua.065@cranfield.ac.uk)}
\address[2]{Cranfield University (e-mail: a.zolotas@cranfield.ac.uk)}
\address[3]{University of Oxford, Cranfield University (e-mail: humd0244@ox.ac.uk, miguel.aranacatania@cranfield.ac.uk)}

\markboth
%{Author \headeretal: Preparation of Papers for IEEE TRANSACTIONS and JOURNALS}
%{Author \headeretal: Preparation of Papers for IEEE TRANSACTIONS and JOURNALS}
{Author }
{Author }

\corresp{Corresponding authors: Miguel Arana-Catania, Argyrios Zolotas. Responsible Corresponding Author: Argyrios Zolotas (e-mail: a.zolotas@cranfield.ac.uk).}

\begin{abstract}
Remote sensing has become a vital tool across sectors such as urban planning, environmental monitoring, and disaster response. While the volume of data generated has increased significantly, traditional vision models are often constrained by the requirement for extensive domain-specific labelled data and their limited ability to understand the context within complex environments. Vision Language Models offer a complementary approach by integrating visual and textual data; however, their application to remote sensing remains underexplored, particularly given their generalist nature. This work investigates the combination of vision models and VLMs to enhance image analysis in remote sensing, with a focus on aircraft detection and scene understanding. The integration of YOLO with VLMs such as LLaVA, ChatGPT, and Gemini aims to achieve more accurate and contextually aware image interpretation. Performance is evaluated on both labelled and unlabelled remote sensing data, as well as degraded image scenarios which are crucial for remote sensing. The findings show an average MAE improvement of 48.46\% across models in the accuracy of aircraft detection and counting, especially in challenging conditions, in both raw and degraded scenarios. A 6.17\% improvement in CLIPScore for comprehensive understanding of remote sensing images is obtained. The proposed approach combining traditional vision models and VLMs paves the way for more advanced and efficient remote sensing image analysis, especially in few-shot learning scenarios.

\end{abstract}


\titlepgskip=-21pt

\maketitle

\section{Introduction}
\label{sec:introduction}
\PARstart{R}{emote} sensing technology has become an indispensable tool in various fields, significantly enhancing capabilities in urban planning, environmental monitoring, and disaster response \cite{navalgund2007remote, boccardo2015remote, patino2013review}. As remote sensing capabilities advance, the volume of data generated continues to grow exponentially, providing invaluable insights for various applications. This surge in data also drives the need for more sophisticated analytical tools to process, interpret, and extract actionable information from the complex datasets generated by remote sensing. Efficiently analysing and interpreting vast amounts of imagery data is crucial for making informed decisions. As remote sensing technology continues to advance, the integration of AI will play
a crucial role in unlocking its full potential \cite{zhang2022artificial}.

Traditional vision models, such as You Only Look Once (YOLO) \cite{jiang2022review}, have demonstrated remarkable performance in object detection and image classification tasks. However, these models often require large amounts of domain-specific labelled data, which can be challenging to obtain in the remote sensing context \cite{wu2023yolo}. Vision Language Models (VLMs) \cite{zhang2021vinvl} have introduced a new dimension to image analysis by combining visual and textual data, enabling tasks like Visual Question Answering (VQA) \cite{antol2015vqa}. These models offer a promising avenue for improving image interpretation, especially in scenarios where labelled data is sparse. Yet, their performance in specialised domains, such as remote sensing, remains uncertain due to their generalist nature \cite{kuckreja2024geochat}. 

This paper aims to bridge existing gaps by exploring the synergistic combination of vision models and VLMs to enhance image analysis in remote sensing applications. The research aims to leverage the strengths of both model types to achieve more accurate, efficient, and contextually aware image interpretation. Specifically, the effectiveness of combining YOLO and VLMs such as LLaVA (Language-augmented Vision Attention) \cite {liu2024visual}, ChatGPT \cite{islam2024gpt}, and Gemini \cite{rane2024gemini}, will be investigated for tasks including VQA and visual captioning within the context of few-shot learning in remote sensing applications. In particular, this paper makes the following contributions:
\begin{itemize}
    \item Conducting quantitative and qualitative analysis of various VLMs’ performance on both labelled and unlabelled remote sensing data in few-shot learning scenarios.
\item Exploring how these models perform under degraded image conditions to simulate real-world remote sensing challenges.
\end{itemize}

\section{Related Work}
The integration of vision models, such as YOLO, with VLMs presents a promising approach to overcoming their individual limitations and leveraging their complementary strengths for enhanced image analysis in remote sensing applications.

The combined use of vision models and VLMs potentially offers several key advantages that significantly enhance the capability of remote sensing image analysis. One of the possible benefits is the improvement in accuracy and interpretability. Vision models are renowned for their precise object detection capabilities, excelling at identifying and localising objects within images. This capability is crucial for tasks such as vehicle detection in traffic monitoring, building identification in urban planning, and tree counting in forest management \cite{zhu2017deep,zhang2016deep}. However, these models often lack the ability to provide a contextual understanding of the detected objects. This is where VLMs come into play, as they contribute a layer of semantic analysis that offers deeper insights into the context and relationships between the detected objects. By integrating vision models with VLMs, the combined system can deliver richer, more comprehensive interpretations of complex scenes within remote sensing data, thereby improving both the accuracy and the meaningfulness of the analysis \cite{zhang2024vision}.

Another possible advantage of integrating vision models with VLMs is the enhanced performance in degraded conditions. Remote sensing frequently involves dealing with challenging environmental conditions such as atmospheric disturbances, sensor noise, and varying light levels, all of which can degrade image quality. Vision models, with their robust feature detection capabilities, are adept at identifying key elements even in suboptimal images. When these capabilities are combined with the contextual inference abilities of VLMs, the system becomes more resilient to noise and image artifacts. VLMs can infer context and meaning even when the visual data is partially obscured or noisy, leading to more reliable and consistent performance in real-world remote sensing applications \cite{rasti2021image}.

Moreover, the combination of vision models and VLMs addresses the challenge of data scarcity, which is often encountered in remote sensing. Acquiring large labelled datasets for training vision models can be difficult, particularly in specialised or less-studied environments where labelled data is limited. VLMs, however, have the ability to perform effectively even with fewer labelled examples, making them a valuable complement to vision models. By leveraging VLMs in conjunction with vision models, it becomes possible to develop a more efficient approach that maximises the utility of available data. This is particularly beneficial in remote sensing contexts where data scarcity is a common challenge \cite{meng2024few}, allowing for the creation of robust models that can operate effectively even with limited labelled datasets.

From a methodological perspective, existing work on integrating vision models with VLMs can be broadly grouped into three strategies: (a) direct prompting of VLMs with raw images, (b) pre-processing with detection or segmentation models before VLM reasoning (the approach adopted in this study), and (c) joint multimodal training where both visual and language components are optimised together. Strategy (a) offers simplicity but can suffer from irrelevant attention in cluttered scenes; strategy (b) provides strong visual grounding while retaining flexibility in the language stage; and strategy (c) yields tight integration but often requires large, domain-specific datasets.

In terms of VLM architectures, prior studies span CLIP-based encoders, instruction-tuned multimodal LLMs (e.g., LLaVA), and proprietary foundation models (e.g., GPT-4o, Gemini). CLIP-based models are efficient and strong in zero-shot transfer but may lack rich reasoning capabilities. Instruction-tuned models improve task adaptability but incur higher inference costs. Proprietary models generally offer the most robust reasoning and degraded-image performance but may face deployment constraints due to their closed-source nature and computational requirements.

Application domains include fine-grained object detection, activity recognition, land cover classification, and environmental monitoring. Detection-focused studies (e.g., traffic monitoring, infrastructure mapping) often favour pre-processing approaches, while scene-level interpretation tasks (e.g., disaster response, urban planning) tend to rely on direct prompting or joint training. Our work applies the pre-processing approach to two distinct remote sensing contexts—structured counting (aircraft detection) and unstructured interpretation (disaster imagery)—demonstrating adaptability across different domain demands.

Several studies, such as \cite{saeedbeyond, samuel2024integrating, shijun2024monitoring, tan2024multi}, have explored the combination of vision models and VLMs. They have explored combining vision models with VLMs such as Faster R-CNN, YOLO, ChatGPT, and LLaVA to address challenges in various fields, including steel surface defect detection, real-time water pollution surveillance, and navigation.  Two of these studies \cite{saeedbeyond, shijun2024monitoring} are explained in more detail below.

\subsection{Case Study: Distance-Aware Large Vision Models for Smarter
Navigation}

Saeed et al. \cite {saeedbeyond} explore the enhancement of autonomous vehicle navigation by integrating Large Vision Models with advanced computer vision techniques to improve spatial awareness and distance estimation. They highlight the critical role of accurate distance measurement in ensuring safer and more efficient driving. The study investigates methods such as stereo vision, monocular depth estimation, and triangle similarity to estimate distances between vehicles and objects within the driving environment. Additionally, the research incorporates the LLAVA natural language processing model to generate detailed scene descriptions by combining object detection data with distance measurements, thereby improving situational awareness. This work is closely related to our work, as it similarly focuses on integrating advanced computer vision techniques with VLMs to enhance scene understanding and object detection. Although the application areas differ—autonomous driving versus remote sensing—the underlying objectives of improving visual data accuracy and contextual interpretation are aligned.

\subsection{Case Study: Monitoring Human Activities with YOLO and LLaVA}

A notable example of the successful integration of a vision model and a VLM is presented in the study by Shijun Pan et al. \cite{shijun2024monitoring}, where YOLOv8 was combined with LLaVA to monitor human activities along the Asahi River in Okayama Prefecture, Japan. The objective was to automate the recognition of human activities from 4K camera images using YOLO for object detection and LLaVA for activity analysis. The study successfully demonstrated that this combination could effectively count people and identify activities such as walking, running, and skateboarding. By automating these tasks, the approach addressed the limitations of manual surveys and provided detailed insights into how the river space was utilised. However, challenges were noted in detecting activities at longer distances and in different spatial contexts, suggesting areas for further refinement. This case study highlights the potential of combining object detection models with VLMs to enhance the analysis of dynamic scenes, a capability that can be extended to various remote sensing applications. 

\section{Methodology}
Our methodology combines the advanced object detection capabilities of YOLOv8 with the interpretative power of VLMs such as ChatGPT, LLaVA and Gemini to create a system capable of both identifying and describing objects in complex visual datasets. %See Figure \ref{fig:Methodology Overview} for the methodology overview. 
The project centres on the Airbus Aircraft Detection Dataset \cite{airbusgeo2023}, a specialised collection of annotated images designed for the task of aircraft detection. The Airbus Aircraft Detection Dataset contains 103 high-resolution images. Alongside the primary aircraft dataset, another dataset with selected disaster satellite images \cite{Hajkova2024} is being utilised for qualitative investigation. 

The workflow begins with data preparation, where the Airbus dataset is enhanced through a series of modifications, including data augmentation techniques such as rotation, scaling, and flipping. These adjustments are designed to enhance the diversity of the training data, thereby increasing the model’s robustness. A separate set of degraded images, adding Gaussian noise, is also prepared. 

Once the data is prepared, YOLOv8, a cutting-edge object detection model known for its speed and accuracy, is fine-tuned on this dataset. The model learns to accurately detect and localise aircraft within the images, outputting precise bounding boxes around each detected object. The outputs from YOLOv8 are then integrated with VLMs.

The methodology incorporates an evaluation framework that assesses the performance of the integrated model using both quantitative and qualitative metrics, specifically through VQA and visual captioning tasks for both raw and labelled data. A quantitative metric, Mean Absolute Error (MAE),  provides a detailed analysis of the object detection accuracy. To ensure the model's robustness and applicability in real-world scenarios, it is also tested under degraded conditions. Additionally, a natural language processing metric, CLIPScore \cite{hessel2021clipscore}, is used to evaluate the quality and relevance of the textual descriptions generated by the VLMs. A manual evaluation of the outputs is also conducted. 

\section{Implementation}
\begin{figure}[htbp]
    \centering
    \begin{subfigure}[b]{1\linewidth}
        \centering
        \includegraphics[width=0.8\linewidth]{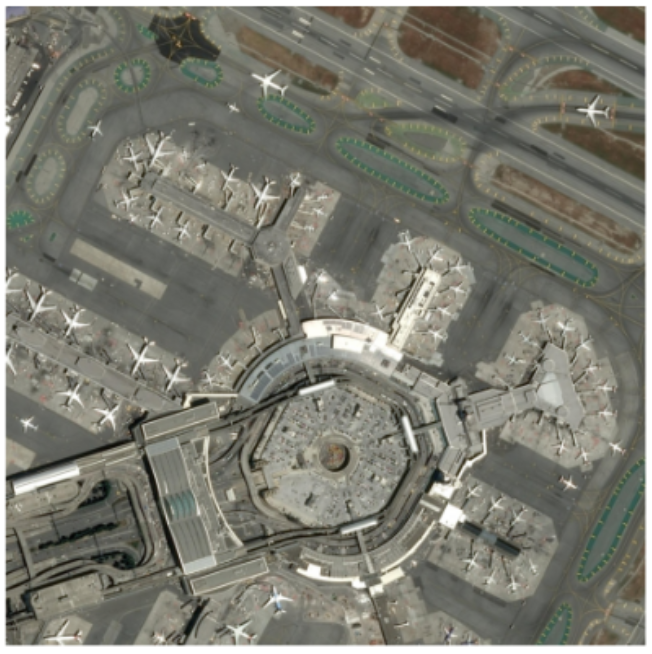}
        \caption{Raw Image}
        \label{fig:raw image}
    \end{subfigure}
    \hfill
    \begin{subfigure}[b]{1\linewidth}
        \centering
        \includegraphics[width=0.8\linewidth]{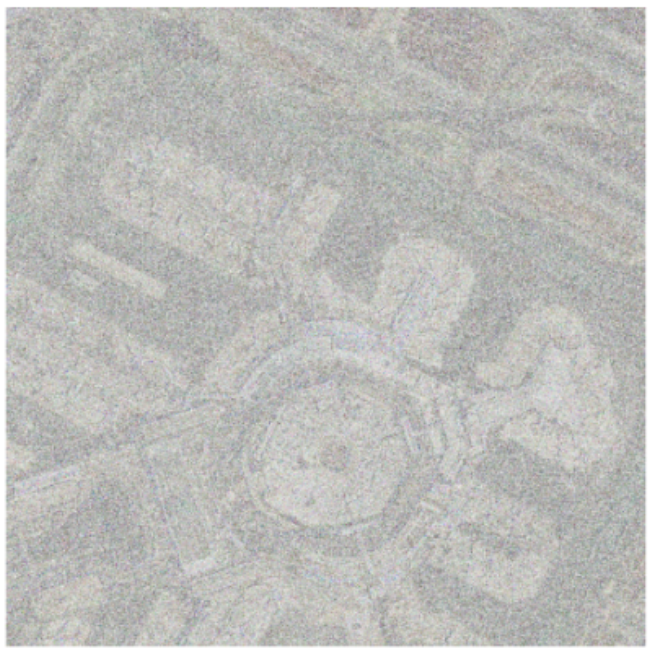}
        \caption{Degraded Image}
        \label{fig:degraded image}
    \end{subfigure}
    \caption{Raw and Degraded Image comparison example}
    \label{fig:comparison_images}
\end{figure}

Our implementation integrates the YOLOv8s model for object detection with VLMs to enhance interpretative analysis. The Airbus Aircraft Detection Dataset, comprising 103 high-resolution remote sensing images, is modified to create both standard and degraded image sets. The dataset is illustrated in Figure \ref{fig:comparison_images}. 
The degradation process simulates challenging real-world conditions by adding Gaussian noise. A noise array is generated using a normal distribution with a mean of 0 and a standard deviation of 50, and subsequently added to the original image, resulting in a noisy version where the pixel values are randomly increased or decreased based on the generated noise value.

YOLOv8s, a smaller variant of YOLOv8, using pretrained weights \cite{yolov8_weights}, is fine-tuned on this dataset with 50 epochs, a batch size of 16, and an image size of 512x512 pixels. The aim of fine-tuning is to optimise both precision and recall, ensuring that the model accurately identifies and locates objects with precise bounding boxes in the dataset. To optimise training efficiency, automatic mixed precision is employed, with a cosine learning rate schedule starting at 0.01. Data augmentation techniques, including random horizontal flipping, colour adjustments, rotation, scaling, and brightness-adjustment, are applied to enhance the model's generalisation. A 5-fold cross-validation approach ensures robust evaluation and mitigates overfitting. Throughout the training process, the model's performance was continuously monitored by evaluating precision and recall on the validation set. This allowed for adjustments to the training strategy, such as refining the loss function or adjusting the training duration, to optimise the results. The fine-tuning process focused on optimising the model's hyperparameters to achieve the best balance between precision and recall.

\begin{figure}[htbp]
    \centering
    \includegraphics[width=0.95\linewidth]{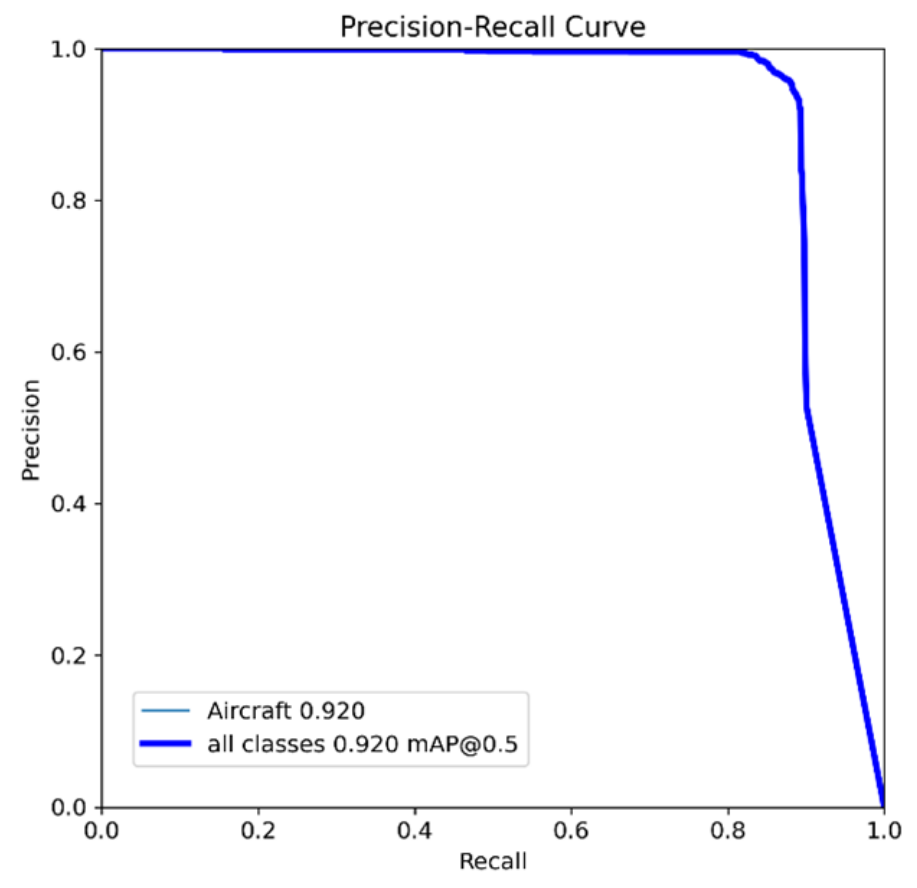}
    \caption{Precision-Recall Graph}
    \label{fig:Precision and Recall Graph}
\end{figure}

As shown in the Precision-Recall Curve in Figure \ref{fig:Precision and Recall Graph}, the final model consistently maintains high precision across a wide range of recall values. This indicates its ability to accurately detect aircraft while minimising false positives. The curve's sharp drop near maximum recall highlights the model's capacity to detect nearly all aircraft instances, with only slight compromises in precision at the highest recall levels. The mean average precision MAP@0.5 value of 0.920 further demonstrates the model’s effectiveness in delivering accurate detections and precise bounding boxes, ensuring reliable performance in real-world applications.

After fine-tuning, YOLOv8s is integrated with the pre-trained VLMs LLaVA, ChatGPT, and Gemini to generate descriptions and answer questions about detected objects. The study employs specifically ChatGPT-4o \cite{openai_gpt4o}, LLaVA v1.5-13B-3GB \cite{llava_v1.5}, and Gemini 1.5 Flash \cite{google_ai}, to perform these tasks. 
ChatGPT-4o integrates a vision encoder with GPT-4’s extensive language model.  LLaVA v1.5 is built on a large-scale transformer architecture with 13 billion parameters. It uses a dual-encoder system, where one encoder processes visual data and the other handles language processing. The model's 3GB version is optimised for efficiency, making it capable of processing large datasets in real-time. Gemini 1.5 Flash leverages the Mixture of Experts (MoE) approach to create a more specialised and efficient
system. The MoE architecture divides the model into smaller, distinct ``expert” neural networks, each trained to handle specific input types.

Examples of the inputs to the pipeline include raw and labelled images in both standard and degraded conditions. %as shown in Figure \ref{fig:comparison_images}. 
The prompts are also illustrated in Table \ref{tab:prompts_vlms}. Evaluation includes both a quantitative metric, mean absolute error (MAE) for object counting, and qualitative assessments, using CLIPScore \cite{hessel2021clipscore} and manual evaluations of description accuracy and relevance. This framework ensures rigorous testing under various conditions, emphasising the models' real-world applicability.

\begin{table}[htbp]
\centering
\caption{Prompts given to VLMs for Different Tasks}
\begin{tabular}{|p{2.2cm}|p{4.9cm}|}
\hline
\textbf{Tasks} & \textbf{Prompts given to VLMs} \\
\hline
\textbf{VQA} & \textbf{Aircraft Dataset}.
Raw: ”How many aircraft are there in this picture?”
Raw + Bounding Boxes: ”How many aircraft are there in
this picture? Use the aircraft detected by YOLO indicated
by the bounding boxes to aid your assessment.” \\
\hline
\textbf{VQA} & \textbf{Disaster Response Dataset}.
Raw: ”Which route is unobstructed?”
Raw + Bounding Boxes: ”Which route is unobstructed?
Use the vehicles detected by YOLO indicated by the bounding boxes to aid your assessment.”\\
\hline
\textbf{Visual Captioning} & Raw: ”What does the image depict?”
Raw + Bounding Boxes: ”What does the image depict? Use
the aircraft detected by YOLO indicated by the bounding
boxes to aid your description.”\\
\hline
\end{tabular}
\label{tab:prompts_vlms}
\end{table}

The evaluation includes three key tasks: 1) VQA on the aircraft dataset for counting aircraft in both standard and degraded images, 2) VQA on the disaster response dataset, and 3) visual captioning on the standard aircraft dataset.

\section{Results}
In this section are presented the quantitative and qualitative performance outcomes from integrating YOLOv8 with LLaVA, ChatGPT, and Gemini.
 
\subsection{Quantitative Results}

The quantitative results are summarised in Table \ref{tab:vlm_summary}, showing the object counting accuracy for each VLM, presenting the MAE for raw images, degraded images, raw images with bounding boxes, and degraded images with bounding boxes. The raw and degraded results represent the baseline methods to which we compare our proposed approach, which is represented by the rows including the bounding boxes. Although YOLOv8s was evaluated in isolation to understand its baseline detection behaviour, detailed YOLO-only results are not reported in this work, as the focus is on how Vision–Language Models can complement detection outputs through contextual reasoning. The YOLO component serves primarily as a region proposal stage, enabling the VLM to operate on candidate areas of interest rather than the full scene. It should also be noted that the rest of the tasks evaluated, besides this counting task, cannot be performed by a standalone YOLO model, since they are not quantitative tasks.

\begin{table}[htbp]
\centering
\caption{MAE Summary for Each VLM under Different Conditions}
\begin{tabular}{|p{3.0cm}|p{1.2cm}|p{1.2cm}|p{1.2cm}|}
\hline
Data & LLaVA & ChatGPT & Gemini \\
\hline
Raw & Undefined\footnotemark & 8.45 & 35 \\
Degraded & Undefined & 17.27 & 103.81 \\
Raw + bounding boxes & 26.09 & 8.27 & 16.27 \\
Degraded + bounding boxes & 26.18 & 8.54 & 10.72 \\
\hline
Improvement (Raw to Raw + bounding boxes) & - & 2.13\% & 53.51\% \\
Improvement (Degraded to Degraded + bounding boxes) & - & 50.54\% & 89.67\% \\
\hline
Average Improvement Across Scenarios & \multicolumn{3}{c|}{48.46\%} \\
\hline
\end{tabular}
\label{tab:vlm_summary}
\end{table}
\footnotetext{
Undefined results indicate that the model is unable to generate a response.}

LLaVA was unable to generate a response (indicated in the Table as "Undefined") under both standard and degraded conditions, but was able to respond when including the bounding boxes. ChatGPT showed moderate performance across both raw and degraded images, with noticeable improvements when bounding boxes were applied, especially under degraded conditions. Gemini demonstrated the most significant improvements, particularly in degraded conditions. 

Overall, the results suggest that the inclusion of bounding boxes generally enhances the performance of all models, particularly in challenging scenarios involving degraded images, with an average improvement of 48.46\% across scenarios. Details will be further discussed in the next section. 
 
\subsection{Qualitative Results}

Both scaled CLIPScore and manual evaluations were investigated, showing qualitative improvements in the results, based on raw images. 

The summary of the CLIPScore outcomes is shown in Table \ref{tab:clipscore_summary}. The raw results represent the baseline methods to which we compare our proposed approach, which is represented by the rows including the bounding boxes. As mentioned in the previous section, CLIPScore may not reliably measure conceptual alignment in degraded images, as image degradation can distort key features, leading to inaccurate assessments that are more influenced by visual quality than underlying content, making raw image comparison more suitable for qualitative assessment in such cases. The results suggest that the inclusion of bounding boxes generally enhances model performance, as evidenced by higher CLIPScores as shown in Table \ref{tab:clipscore_summary}. Bounding boxes likely assist the models by directing their focus to relevant areas of the image, resulting in more accurate and contextually appropriate descriptions. Overall, the CLIPScore results demonstrate that while all models perform moderately with raw images, the addition of bounding boxes improves description accuracy, particularly for ChatGPT and LLaVA, with an average improvement of 6.17\% across models. 

\begin{table}[htbp]
\caption{CLIPScore Summary for Each VLM under Different Conditions}
\centering
\begin{tabular}{|p{3cm}|p{1cm}|p{1cm}|p{1cm}|}
\hline
Data & LLaVA & ChatGPT & Gemini \\
\hline
Raw & 31.88 & 31.88 & 32.39 \\
Raw + bounding boxes & 34.09 & 34.09 & 33.90 \\
\hline
Improvement (\%) & 6.93\% & 6.93\% & 4.66\% \\
\hline
Average Improvement Across Models & \multicolumn{3}{c|}{6.17\%} \\
\hline
\end{tabular}
\label{tab:clipscore_summary}
\end{table}

\begin{figure} [htbp]
    \centering
    \includegraphics[width=0.8\linewidth]{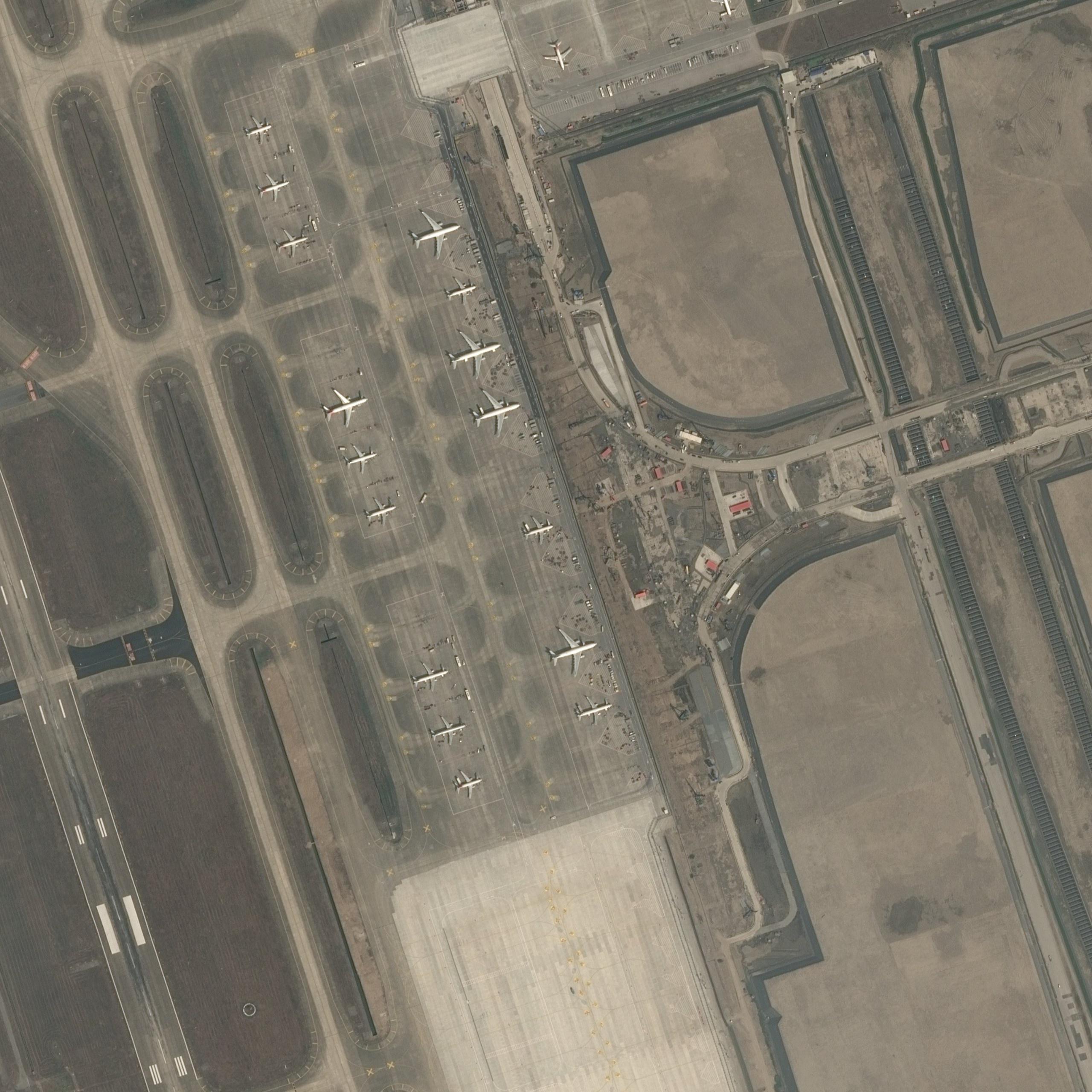}
    \caption{Illustration of different descriptions provided by different VLMs (Example 1)}
    \label{fig:example_of_different_descriptions_by_different_vlm}
\end{figure}

These findings are further corroborated by manual evaluations as shown in Tables \ref{tab:vlm_raw_descriptions} and \ref{tab:vlm_raw_boundingboxes_descriptions} corresponding to Figure \ref{fig:example_of_different_descriptions_by_different_vlm}, which provide additional insights into the relevance and correctness of the generated descriptions, confirming the benefits of bounding box integration. Detailed discussions of these examples will be provided in the next section. %Additional examples can be found in Appendix \ref{appendix}, which provides further insights to support this finding. 
Manual evaluation was also performed on a secondary dataset of disaster satellite imagery \cite{Hajkova2024}. This evaluation revealed the potential for uncovering additional insights across different contexts, as the model was able to deduce better answers when bounding boxes were added. Overall, the bounding boxes in the dataset assist VLMs in estimating the number of aircraft, improving positioning accuracy, and more effectively detecting the presence of other objects in the image description. This will be explored in greater detail in the following section.

\begin{table*}[htbp]
\centering
\caption{Detailed Descriptions provided by Different VLMs for Raw Image Data (Example 1)}
\resizebox{\textwidth}{!}{%
\begin{tabular}{|p{0.7cm}|p{5cm}|p{5cm}|p{5cm}|}
\hline
Data & LLaVA & ChatGPT & Gemini \\
\hline
Raw & 
The image appears to be an aerial or satellite view of an airport. The layout includes multiple runways, taxiways, and aprons where airplanes are parked. There are several aircraft visible, both on the ground and on the runways, indicating that the airport is operational. The image is somewhat blurry, which makes it difficult to discern specific details about the aircraft or the airport's infrastructure. The surrounding area looks like it could be a mix of undeveloped land and possibly some industrial or airport-related facilities. The overall color palette is muted, with a lot of earth tones, which is typical for an airport environment. & 
The image is an aerial view of an airport. Here are some details: Airplanes: Several airplanes are parked in designated areas. The aircraft appear to be commercial airliners, aligned parallel to each other. Runways and Taxiways: There are multiple runways and taxiways visible. The runways are distinct with their long, straight paths and markings. Apron Area: The planes are parked on an apron area with clearly marked lines for guidance. Construction: On the right side of the image, there are areas under construction with various equipment and incomplete structures. Surroundings: The surroundings of the airport include large open spaces, likely fields or unused land. This image likely captures a busy airport with ongoing construction or expansion work. & 
The image you sent me shows an aerial view of a large airport with airplanes parked on the tarmac. The terminal buildings are visible in the background, along with other airport structures. The sky is clear and blue. While the exact location of the airport in the image is difficult to determine, the overall layout is similar to that of Shanghai Pudong International Airport. \\
\hline
\end{tabular}
}
\label{tab:vlm_raw_descriptions}
\end{table*}

\begin{table*}[htbp]
\centering
\caption{Detailed Descriptions provided by Different VLMs for Raw Image Data with Bounding Boxes (Example 1)}
\resizebox{\textwidth}{!}{%
\begin{tabular}{|p{0.7cm}|p{5cm}|p{5cm}|p{5cm}|}
\hline
Data & LLaVA & ChatGPT & Gemini \\
\hline
Raw + Bounding boxes & 
The image appears to be an aerial or satellite view of an airport or airfield. There are multiple aircraft parked on the tarmac, which are likely passenger planes given their size and configuration. The planes are scattered across several aprons or parking areas, which are the paved areas where aircraft are parked, loaded, unloaded, or prepared for flights. The airport has a network of taxiways and runways, which are the paved paths that aircraft use to move around the airport. The runways are the long, straight, and wide areas where planes take off and land. There are also several buildings and structures visible, which could be hangars, terminals, or other airport facilities. The image is taken from a high angle, providing a comprehensive view of the airport layout and operations. The image is marked with several rectangles, which might indicate areas of interest or specific locations within the airport. These markings could be used for planning, analysis, or other purposes related to airport operations or security. The overall condition of the airport seems to be well-maintained, with clear markings and a structured layout. The presence of multiple planes suggests that the airport is active and operational. & 
The image appears to be a satellite view of an airport. The following features can be observed: Runways and Taxiways: Multiple parallel runways and taxiways are visible on the left side of the image. Aircraft: There are numerous aircraft parked on the tarmac. Each aircraft is highlighted with a box around it, indicating annotation. Construction Area: On the right side, there is a large construction area with several buildings and open spaces, indicating ongoing development or expansion. Apron Area: The aircraft are primarily concentrated in the apron area, which is used for parking, loading, and unloading. Paved Areas: The image also shows large paved areas, possibly used for taxiing aircraft or additional parking spaces. The annotated boxes around the aircraft suggest that this image might be used for some form of analysis or monitoring, possibly related to airport operations, aircraft movements, or construction progress. & 
Based on the image you sent, it appears to be an aerial view of a medium-sized commercial airport. Here's a description: In the foreground, there are several airplanes parked on the tarmac. They appear to be a mix of passenger airplanes and smaller cargo planes. There are long rectangular buildings in the background that are likely the terminal buildings. Adjacent to the terminal buildings, there are other airport buildings and structures, including what appears to be a control tower. The sky is clear and bright blue. \\
\hline
\end{tabular}
}
\label{tab:vlm_raw_boundingboxes_descriptions}
\end{table*}

\subsection{Computational performance}

To assess the practical feasibility of the proposed pipeline, we measured its empirical runtime under the evaluation setup, using an NVIDIA A100 GPU for YOLOv8s inference and either local or API-based queries for VLM processing. YOLOv8s detection averaged $\sim$12 ms per 512×512 image. VLM processing per detected bounding box required $\sim$300 ms using LLaVA (local), $\sim$0.9 s using Gemini 1.5 Flash (API), and $\sim$1.2 s using ChatGPT-4o (API). For typical scenes in our dataset, the total processing time ranged from $\sim$1.0–1.5 s (YOLO + LLaVA) to $\sim$3–6 s (YOLO + API-based VLMs). These timings indicate that the approach is well suited for batch or offline remote sensing analysis, while real-time deployment would benefit from optimisation techniques such as bounding box prioritisation or local VLM hosting.

\section{Discussion}

This section presents a rigorous discussion of performance for the results presented earlier. Detailed interpretations and potential implications are provided. The performance gains observed with the YOLO + VLM pipeline over VLM-alone appear to arise from the complementary nature of detection and multimodal reasoning rather than from raw detection accuracy. First, YOLO-generated bounding boxes could narrow the VLM’s visual focus to relevant regions, suppressing background clutter that is common in remote sensing imagery. Second, this spatial grounding could reinforce contextual understanding, enabling VLMs to maintain consistency in counting and scene interpretation even when inputs are degraded. Third, the two components could lead to error-compensating behaviour: VLMs could infer missed detections by recognising spatial patterns or repetitive arrangements (e.g., partially occluded rows of aircraft), while YOLO’s detections could constrain the VLM from overcounting due to hallucination or misinterpretation. 

We have observed in the results how YOLO occasionally failed to detect partially occluded or overlapping aircraft, resulting in undercounts. The VLM, informed by the bounding boxes of detected objects and broader scene context, was able to infer the presence of missing aircraft. On the other hand, in some cluttered airport scenes, YOLO misclassified ground vehicles or structures as aircraft. The VLM, when reasoning over bounding box crops, discounted these false positives, bringing the count closer to the ground truth. 

\subsection{Object Detection}
The Mean Absolute Error (MAE) quantifies the accuracy of a model’s predictions by calculating the average absolute difference between the predicted and actual object counts. In the context of the provided results shown in Table \ref{tab:vlm_summary}, the MAE values reveal significant variations in performance across different VLMs and under various conditions.

\begin{figure} [htbp]
    \centering
    \includegraphics[width=0.93\linewidth]{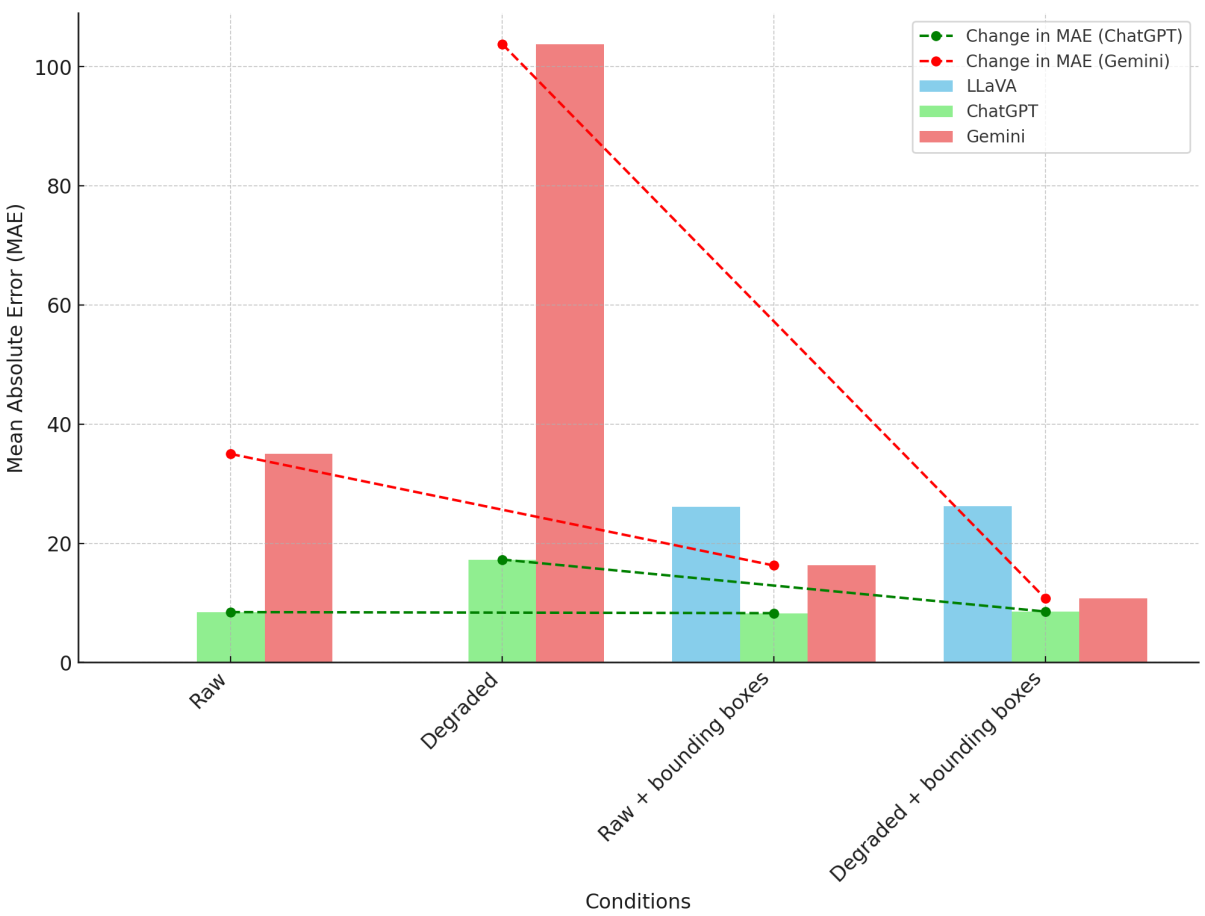}
    \caption{MAE Improvement over VLMs}
    \label{fig:MAE improvement over VLMs}
\end{figure}

These findings, as depicted in Figure \ref{fig:MAE improvement over VLMs}, underscore the critical role of bounding boxes in enhancing the accuracy of object counting, particularly in challenging visual environments. The downward trend in these lines signifies a reduction in MAE. This reduction indicates that the models perform better (i.e., more accurately) when the data is supplemented with bounding boxes. The significant reduction in MAE for Gemini and the noticeable improvement for LLaVA illustrate the effectiveness of bounding boxes in providing essential contextual clues, leading to more accurate and reliable model performance.

\subsubsection{Raw and Degraded Conditions}

On raw images, ChatGPT performs well with an MAE of 8.45, indicating a high level of accuracy.
When the images are degraded, the MAE increases to 17.27. This increase suggests that ChatGPT’s accuracy is somewhat affected by the lower image quality, though it still manages to maintain reasonable performance.

For raw images, Gemini has an MAE of 35, indicating that it is less accurate compared to ChatGPT in this scenario. The MAE for Gemini rises significantly to 103.81 when the images are degraded, indicating a substantial drop in performance. This suggests that Gemini is more sensitive to the degradation of image quality, which greatly impacts its ability to make accurate predictions.

LLaVA does not return a usable MAE for either raw or degraded images, indicating that the model struggles to generate responses under these conditions. This lack of usable output suggests that LLaVA may require additional context or enhancements, like bounding boxes, to function effectively in these scenarios.

\subsubsection{Raw Images with Bounding Boxes}

With bounding boxes added to raw images, ChatGPT’s MAE improves slightly to 8.27. This improvement, though modest, indicates that even in the best-case scenario (with raw images), the model can further refine its accuracy when given additional contextual information through bounding boxes.

For Gemini, the introduction of bounding boxes to raw images leads to a significant improvement, reducing the MAE to 16.27 from 35. This reduction shows that Gemini benefits substantially from the added context, suggesting that the model can leverage bounding boxes to enhance its performance consistently.

In the case of LLaVA, adding bounding boxes to raw images results in an MAE of 26.09. While this is a usable result, it is still relatively high compared to the other models, indicating that LLaVA may not fully capitalise on the additional context provided by the bounding boxes, even in optimal conditions with raw images.
\subsubsection{Degraded Images with Bounding Boxes}

When bounding boxes are added to the degraded images, ChatGPT’s MAE decreases slightly to 8.54 from 17.27. This indicates that the bounding boxes help the model significantly recover its accuracy, nearly matching its performance on raw images. ChatGPT demonstrates strong capability in utilising bounding boxes to overcome the challenges of degraded images.

Gemini shows a dramatic improvement when bounding boxes are added to the degraded images, with its MAE dropping from 103.81 to 10.72. This substantial reduction highlights how crucial the additional context provided by bounding boxes is for Gemini, allowing it to improve its performance drastically even when dealing with lower-quality images.

LLaVA shows an MAE of 26.18 when bounding boxes are used with degraded images, which is a stable but higher value compared to ChatGPT and Gemini. While the addition of bounding boxes helps LLaVA respond where it couldn't before, it still faces challenges in achieving lower MAE values, indicating that more context or further tuning might be necessary for better performance.

The quantitative results in Table 2 reinforce the observed differences in model reliance on YOLO detections. LLaVA’s MAE changed only marginally with the addition of YOLO bounding boxes (Raw: 8.45 → 8.27; Degraded: 17.27 → 8.54), indicating strong adherence to YOLO’s outputs and limited independent correction. ChatGPT-4o demonstrated moderate reliance, with modest improvement in raw imagery (26.09 → 8.27) but a substantial gain under degraded conditions (35.00 → 16.27), suggesting occasional re-evaluation of detections using scene context. Gemini 1.5 Flash showed the highest degree of independence, producing the largest improvements in both raw (35.00 → 16.27) and degraded imagery (103.81 → 10.72) by actively identifying and correcting YOLO’s overcounting and undercounting. These patterns confirm that variation in “trust level” across VLMs directly influences the degree of benefit derived from integrating YOLO-generated bounding boxes.

\subsection{Visual Captioning}

\begin{figure} [htbp]
    \centering
    \includegraphics[width=0.99\linewidth]{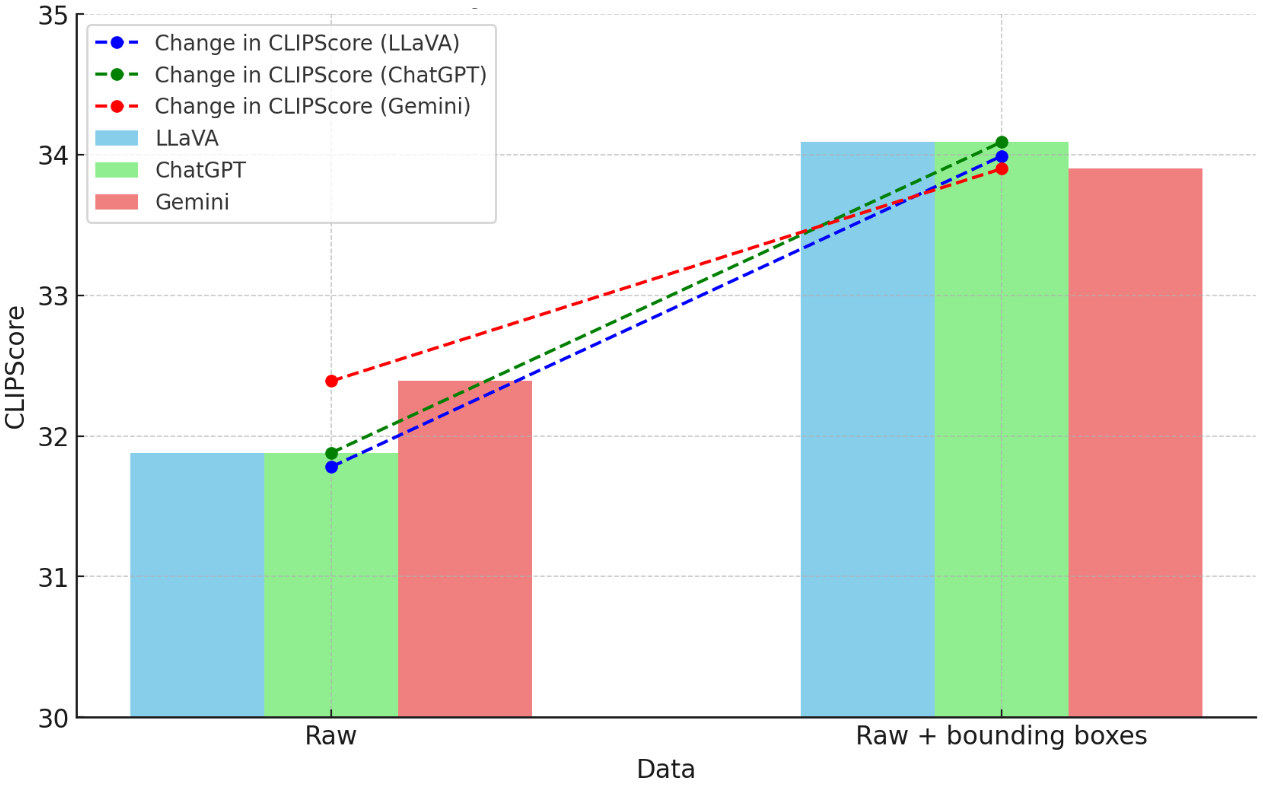}
    \caption{CLIPScore Improvement}
    \label{fig:Clipscore improvement}
\end{figure}

Figure \ref{fig:Clipscore improvement} visualises the improvement in CLIPScore for each VLM when bounding boxes are introduced, as it was shown in Table \ref{tab:clipscore_summary}, offering insights into how well each model aligns image content with the textual descriptions it generates.

For raw images, both LLaVA and ChatGPT achieve a CLIPScore of 31.88, while Gemini slightly outperforms them with a score of 32.39. This indicates that all three models perform similarly in generating descriptions that align with the visual content, with Gemini having a marginal edge. These scores suggest that the models are performing similarly at capturing key elements in the images, such as identifying significant objects like airplanes in an airport setting. The slight variation in scores reflects small differences in how each model processes and interprets raw visual data.

When bounding boxes are introduced, there is a noticeable improvement in CLIPScore for all three models. LLaVA and ChatGPT both see their scores increase to 34.09, a rise of 2.21. This improvement underscores the advantage provided by the additional context that bounding boxes offer, enabling these models to generate more accurate and contextually relevant descriptions. Gemini also shows an improvement, with its CLIPScore rising by 1.51 to 33.90, indicating that it, too, benefits from the visual cues provided by bounding boxes, though slightly less so than LLaVA and ChatGPT.

These results underscore the crucial role of contextual information in improving the descriptive accuracy of VLMs. The consistent performance gains across all models with the addition of bounding boxes indicate that raw images can achieve better alignment with visual content when given extra context. The modest increases in scores suggest that these models may require further fine-tuning, but they demonstrate that incorporating visual cues, such as bounding boxes, effectively enhances their ability to generate more accurate and contextually rich descriptions. Overall, these results, though marginal, still demonstrate that while all models are effective at describing images, their ability to capture additional details and context can be potentially enhanced by the availability of visual cues such as bounding boxes.

\subsection{Contextual Understanding}

The descriptions provided by VLMs in the examples highlight varying levels of contextual understanding as shown in Tables \ref{tab:vlm_raw_descriptions} and \ref{tab:vlm_raw_boundingboxes_descriptions}. In the raw image descriptions, ChatGPT stands out by offering a more detailed and structured interpretation, identifying specific elements like the alignment of airplanes, construction areas, and overall airport activity. When bounding boxes are introduced, all models improve their contextual awareness, with LLaVA recognising the purpose of the annotations for analysis, and ChatGPT further enhancing its description by interpreting the function of different areas within the image.

\subsubsection{Aircraft Related Dataset}

\begin{figure} [htbp]
    \centering
    \includegraphics[width=0.8\linewidth]{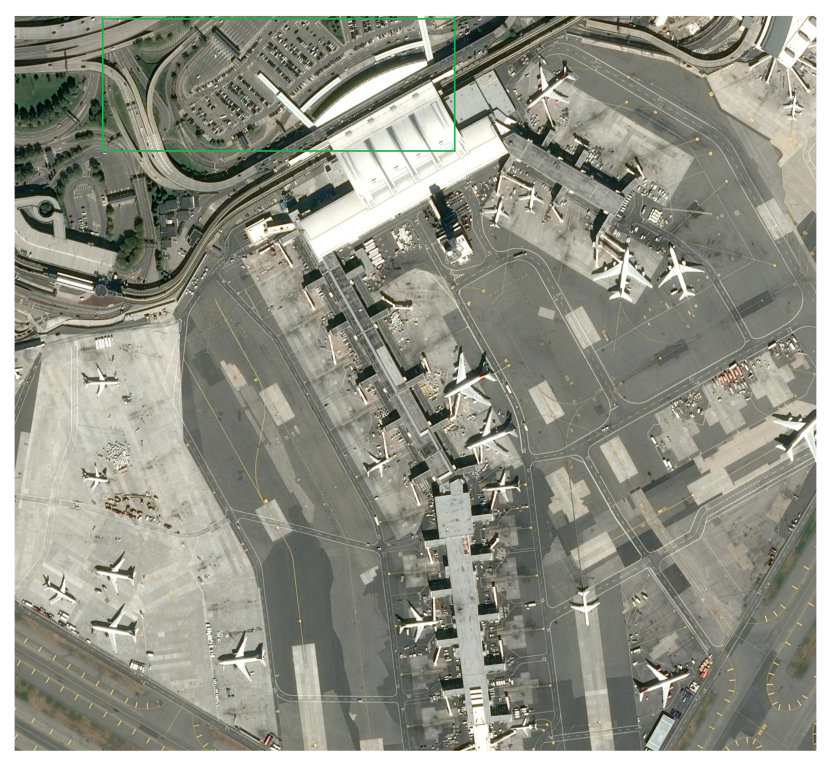}
    \caption{Illustration of Contextual Understanding with Identification of Trucks (Example 2)}
    \label{fig:Example of contextual understanding}
\end{figure}

To show specific examples, the description based on raw images generated by LLavA shown in Tables \ref{tab:comparison_llava_description} and \ref{tab:descriptions} provides an initial understanding of the image, highlighting key elements like the airport layout, multiple runways, taxiways, and aircraft parking stands. It notes the presence of airplanes and suggests that the image is likely an aerial or satellite view of a busy airport. However, the additional details provided through bounding boxes offer a more nuanced understanding. The raw descriptions represent the baseline methods to which we compare our proposed approach, which is represented by the rows including the bounding boxes. As shown in Figure \ref{fig:Example of contextual understanding} and described in Table \ref{tab:comparison_llava_description}, the bounding boxes help to identify the specific number of airplanes —at least 13— positioned at various distances within the image, thereby enhancing the viewer’s awareness of the airport’s scale and activity level. Moreover, the inclusion of trucks and support vehicles near the airplanes, as shown in Figure \ref{fig:Example of contextual understanding}, adds further context, indicating the dynamic operations taking place on the tarmac, such as preparations for takeoff or post-landing procedures. These bounding boxes help in refining the interpretation of the scene, providing a clearer picture of the airport's bustling environment and the intricate activities associated with air travel operations, which were successfully identified by the model as shown in Table \ref{tab:comparison_llava_description}. Similarly, as shown in Table \ref{tab:descriptions} in another example, 
the model also successfully identified at least 14 airplanes in the scene, with some positioned in the foreground and others further in the background, in contrast to the raw image without bounding boxes. This richer context allows for a more precise and informed analysis of the image, emphasising the importance of detailed visual annotations in understanding complex scenes. %Beyond these notable examples that were discussed, additional details on the results can be found in Appendix \ref{appendix}, which provides further insights to support this finding.

\begin{table}[htbp]
\centering
\caption{Comparison of Descriptions generated by LLavA with and without Bounding Boxes (Example 2)}
\begin{tabular}{|p{1.3cm}|p{6cm}|}
\hline
\textbf{Description Type} & \textbf{Description} \\ \hline
\textbf{Raw Description} & The image appears to be an aerial or satellite view of an airport. The layout includes multiple runways, taxiways, and aircraft parking stands. There are several airplanes parked at the gates, and some are on the tarmac, possibly taxiing or preparing for departure. The airport infrastructure includes buildings that could be terminals, hangars, or other airport facilities. The surrounding area shows roads and other transportation infrastructure, indicating that the airport is likely located near a city or a major transportation hub. \\ \hline
\textbf{With Bounding Boxes} & The image is an aerial view of an airport, showcasing a large number of airplanes parked on the tarmac. There are \textbf{at least 13 airplanes visible in the scene}, with some positioned closer to the foreground and others further back. The airplanes are of various sizes and orientations, indicating a busy and active airport. 

In addition to the airplanes, there are \textbf{several trucks scattered around the tarmac, likely serving as support vehicles for the aircraft}. These trucks are positioned near the airplanes, suggesting that they are either preparing for takeoff or have recently landed. The overall scene depicts a bustling airport with numerous airplanes and support vehicles in operation.
\\ \hline
\end{tabular}
\label{tab:comparison_llava_description}
\end{table}

\begin{table}[htbp]
\centering
\caption{Comparison of Descriptions generated by LLavA with and without Bounding Boxes (Example 3)}
\begin{tabular}
{|p{1.3cm}|p{6cm}|}
\hline
\textbf{Description Type} & \textbf{Description} \\ \hline
\textbf{Raw Description} & The image appears to be an aerial or satellite view of an airport. The layout is complex, with multiple runways, taxiways, and aircraft parking stands. There are several airplanes parked at the gates.
\\ \hline
\textbf{With Bounding Boxes} & The image is an aerial view of an airport, showcasing a large number of airplanes parked on the tarmac. There are \textbf{at least 14 airplanes visible in the scene, with some positioned closer to the foreground and others further back}. The airplanes vary in size and are spread across the airport grounds. The airport appears to be quite busy, with numerous planes either preparing for takeoff or having recently landed. \\ \hline
\end{tabular}
\label{tab:descriptions}
\end{table}

The qualitative evaluation of the model's performance goes beyond simple detection and extends to providing additional contextual information. For instance, the presence of trucks at an airport scene can be highlighted using bounding boxes, adding a layer of understanding about the scene's environment. This capability is particularly beneficial in scenarios where context matters, such as distinguishing between different operational areas of an airport or understanding the level of activity in a given region. The ability to annotate and identify multiple objects within a scene enhances the descriptive power of the visual-language models, allowing for more comprehensive and informative descriptions. This additional context not only improves situational awareness but also assists in better decision-making processes by providing a fuller picture of the scene. However, it's important to note that the additional contextual understanding provided by bounding boxes only leads to improvements in some scenarios, rather than consistently enhancing descriptive accuracy across all situations.

\subsubsection{Disaster Related Dataset}

Exploring the secondary dataset of disaster satellite imagery further solidifies that a combined approach enhances situational awareness and supports better decision-making. Upon manual evaluation, integrating bounding boxes with VLMs greatly improves the accuracy in determining the status of evacuation routes, successfully identifying unobstructed paths that the raw images without bounding boxes failed to detect. The bounding boxes, which are generated by YOLO, provide precise localisation of key objects, in this case, vehicles. This spatial information is critical in assessing the condition of the routes.
While object detection models such as YOLO excel at identifying and marking objects within an image, they do not inherently provide a holistic understanding of the scene or context. On the other hand, VLMs are designed to interpret and generate natural language descriptions from visual data but may sometimes lack precision in spatial reasoning or the ability to correlate scattered objects effectively.
By combining the strengths of both YOLO’s object detection and the interpretative capabilities of VLMs, the analysis becomes more robust. The bounding boxes serve as a guide, helping the VLM to better understand the spatial relationships between vehicles and the road infrastructure. This synergy allows for a more accurate assessment of the evacuation routes.

\begin{figure}[htbp]
    \centering
    \begin{subfigure}[b]{1\linewidth}
        \centering
        \includegraphics[width=0.8\linewidth]{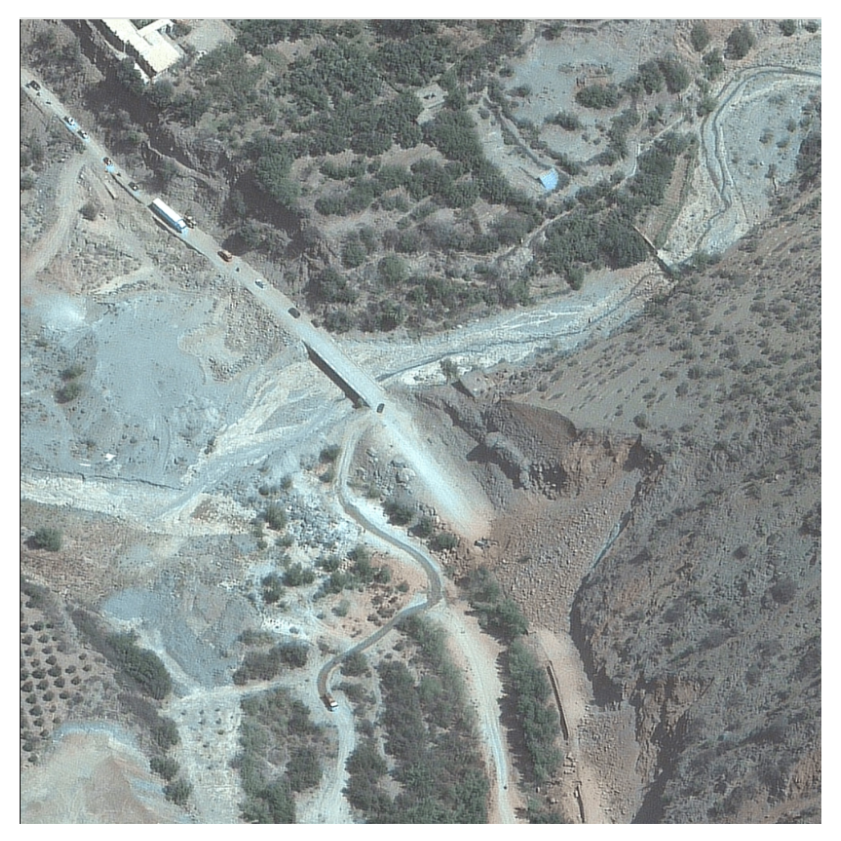}
        \caption{Raw Image of Disaster Area}
        \label{fig:Raw image of disaster area}
    \end{subfigure}
    \hfill
    \begin{subfigure}[b]{1\linewidth}
        \centering
        \includegraphics[width=0.8\linewidth]{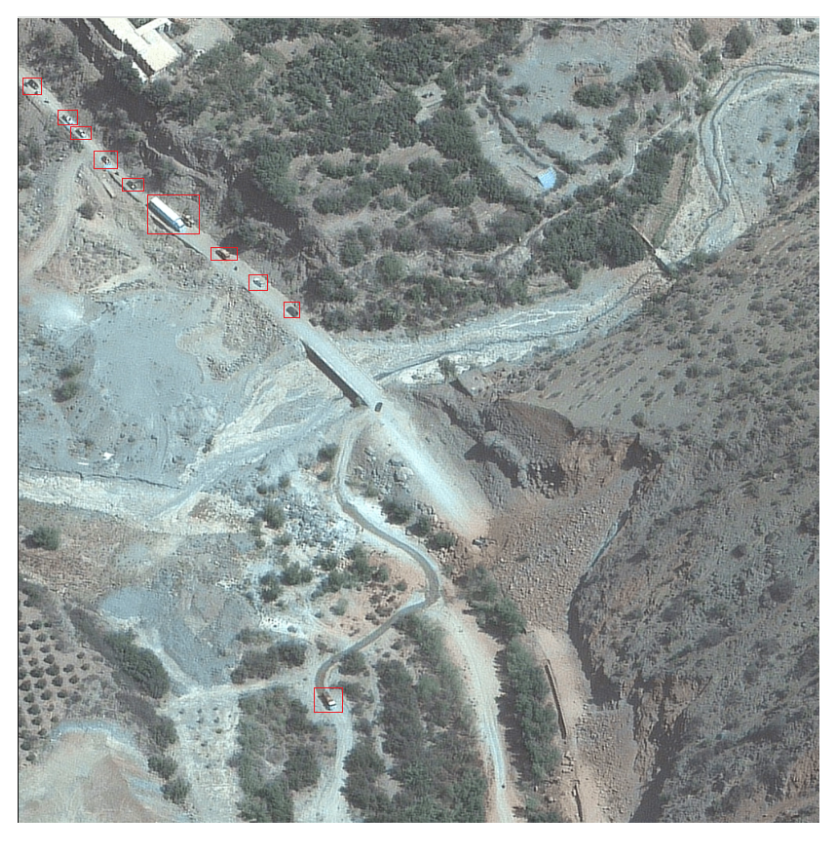}
        \caption{Labelled Image of Disaster Area with Vehicle Detection}
        \label{fig:Labelled(vehicle detection) image of disaster area}
    \end{subfigure}
    \caption{Comparison between Raw and Labelled Images of Disaster Area (Example 4)}
    \label{fig:comparison_disaster_area}
\end{figure}

As seen in Figure \ref{fig:comparison_disaster_area} and Table \ref{table:road_detection}, the bounding boxes highlight vehicles on the primary road, particularly on the approach to the bridge. This visual cue allows the VLM to infer that there is potential congestion, indicating that the route may be partially obstructed rather than fully clear. Similarly, the detection of a single vehicle on the lower road without any surrounding congestion allows the VLM to conclude that this secondary route remains unobstructed.
This combined approach bridges the gap between pure object detection and contextual understanding, leading to a more accurate and actionable assessment, which is crucial for directing relief efforts efficiently. By leveraging both the precise localisation from YOLO and the contextual reasoning of VLMs, the analysis provides a more reliable determination of which evacuation routes are viable, ultimately supporting better decision-making in emergency situations.

\begin{table}[htbp]
\centering
\caption{Road Detection and Assessment Based on Satellite Imagery using ChatGPT (Example 4)}
\begin{tabular}{|p{1.3cm}|p{6cm}|}
\hline
\textbf{Data} & \textbf{Response} \\ \hline
Raw & 
From the image, it appears that there is a primary road crossing a bridge. The road seems to be the main route for vehicles, and there are no visible major obstructions on this road itself. However, there is a section near the bridge where the terrain changes significantly, which could potentially cause issues, but it doesn't seem to fully obstruct the road.

The main road that crosses the bridge and continues appears unobstructed. There are also some minor roads branching off from this main road. These minor roads do not show any clear obstructions either, but their condition might vary due to the terrain.

\textbf{In summary: The primary evacuation route (the main road crossing the bridge) appears to be unobstructed. Minor roads branching off also appear clear, but they should be assessed carefully due to terrain conditions.}
\\ \hline
Raw + Bounding Boxes & 
Based on the updated image with the red detection boxes highlighting vehicles, here is an assessment of the evacuation routes:

Primary Route (Main Road over the Bridge): The main road crossing the bridge is populated with vehicles, indicating that this route is actively being used. However, there are a few vehicles stopped or slowed down on this road, particularly on the approach to the bridge from the top left. This suggests potential congestion or minor obstructions, but the road itself does not appear to be completely blocked. Therefore, this route is partially obstructed.

Secondary Route (Lower Road Leading Away from the Bridge): There is a vehicle detected on the lower road, but it does not seem to be obstructed by any other vehicles or blockages. This route appears to be unobstructed.

\textbf{Summary: 
Primary Route (Main Road over the Bridge): Partially obstructed. 
Secondary Route (Lower Road): Unobstructed and clear for use.}
\\ \hline
\end{tabular}
\label{table:road_detection}
\end{table}

\section{Conclusion}
The work in this paper illustrates that integrating YOLOv8 with VLMs, i.e. LLaVA, ChatGPT, and Gemini, provides a robust framework for aircraft detection and scene interpretation. The combined approach demonstrates the potential for enhancing both the qualitative and quantitative aspects of image analysis. 

The enhancement provided is twofold. From a quantitative viewpoint, it enhances the performance related to object counting, particularly under degraded conditions, with an average MAE improvement of 48.46\% across models in both raw and degraded scenarios. While qualitatively, the use of bounding boxes provides richer contextual insights across multiple objects in various scenarios, as demonstrated in the examples, and leads to a c.6.2\% average improvement in CLIPScore. This enables a more accurate and comprehensive understanding of complex visual data, as demonstrated by notable examples in this paper where the models effectively provided detailed context regarding the number and location of aircraft, identified additional elements within the dataset, and deduced unobstructed routes in disaster scenarios. 

The proposed integration of vision models and VLMs greatly enhances detection accuracy and descriptive quality. It constitutes a robust tool for analysing remote sensing images, particularly for the case of data-scarce environments where zero-shot or few-shot learning is crucial. While this work focused on aircraft detection and disaster response, the methodology is task-agnostic and can be extended to other domains, including land cover classification and change detection, with minimal architectural modifications.

\section*{Acknowledgment}
The authors thank the QinetiQ Fellow and Academic schemes for funding Dr Jeremy Baxter and Dr Richard Lane to initiate the idea and provide supervision during the project.

\bibliography{CUCitations}

\EOD
\end{document}